\documentclass[review]{elsarticle}

\usepackage{lipsum}
\makeatletter
\def\ps@pprintTitle{%
	\let\@oddhead\@empty
	\let\@evenhead\@empty
	\def\@oddfoot{}%
	\let\@evenfoot\@oddfoot} 

\usepackage{graphicx}
\usepackage{enumitem}
\usepackage{siunitx}
\usepackage{amsmath}
\usepackage{multicol}
\usepackage{amssymb}
\usepackage{amsfonts}
\usepackage{float}
\usepackage{natbib}
\usepackage{algorithm}
\usepackage{algorithmic}
\usepackage{array}

\journal{}

\begin{document}

\begin{frontmatter}

\title{Sparse-then-Dense Alignment based 3D Map Reconstruction Method for Endoscopic Capsule Robots}


\author[address1,address2]{Mehmet Turan}
\ead{turan@is.mpg.de}

\author[address4]{Yusuf Yigit Pilavci}
\ead{yigit.pilavci@metu.edu.tr}

\author[address5]{Ipek Ganiyusufoglu}
\ead{gipek@sabanciuniv.edu}

\author[address3]{Helder Araujo}
\ead{helder@isr.uc.pt}

\author[address2]{Ender Konukoglu}
\ead{ender.konukoglu@vision.ee.ethz.ch}

\author[address1]{Metin Sitti}
\ead{sitti@is.mpg.de}


\address[address1]{Max Planck Institute for Intelligent Systems, Stuttgart, Germany}
\address[address2]{Computer Vision Laboratory, ETH Zurich, Switzerland}
\address[address3]{Robotics Laboratory, University of Coimbra, Portugal}
\address[address4]{Electrical and Electronics Engineering, Middle East Technical University, Turkey}
\address[address5]{Computer Science, Faculty of Engineering and Natural Sciences, Sabanci University, Turkey}

\begin{abstract}
In the gastrointestinal (GI) tract endoscopy field, ingestible wireless
capsule endoscopy is emerging as a novel, minimally invasive diagnostic technology
for inspection of the GI tract and diagnosis of a wide range of diseases and
pathologies. Since the development of this technology, medical device companies
and many research groups have made substantial progress in converting passive
capsule endoscopes to robotic active capsule endoscopes which can be controlled by the doctor. However, robotic capsule endoscopy still has some challenges. In particular, the use of such devices to generate a precise and globally consistent
three-dimensional (3D) map of the entire inner organ remains an unsolved
problem. Such global 3D maps of inner organs would help doctors to detect the
location and size of diseased areas more accurately, precisely, and intuitively, thus
permitting more accurate and intuitive diagnoses. The proposed 3D reconstruction system is built in a modular fashion including preprocessing, frame stitching, and shading-based
3D reconstruction modules. We propose an efficient scheme to automatically select the key frames out of the huge quantity of raw endoscopic images. Together with a bundle fusion approach that aligns all the selected key frames jointly in a globally consistent way, a significant improvement of the mosaic and 3D map accuracy was reached. To the best of our knowledge, this framework is the first complete pipeline for an endoscopic capsule robot based 3D map reconstruction containing all of the necessary steps for a reliable and accurate endoscopic 3D map. For the qualitative evaluations, a real pig stomach is employed. Moreover, for the first time in literature, a detailed and comprehensive quantitative analysis of each proposed pipeline modules is performed using a non-rigid esophagus gastro duodenoscopy simulator, four different endoscopic cameras, a magnetically activated soft capsule robot (MASCE), a sub-millimeter precise optical motion tracker and a fine-scale 3D optical scanner.

\end{abstract}

\end{frontmatter}


\section{INTRODUCTION}
Many diseases necessitate access to the internal anatomy of the patient for diagnosis and treatment. Since direct access to most anatomical regions of interest is traumatic, and sometimes impossible, endoscopic cameras have become a common method for viewing the anatomical structure. In particular, capsule endoscopy has emerged as a promising new technology for minimally invasive diagnosis and treatment of gastrointestinal (GI) tract disease.  The low invasiveness and high potential of this technology has led to substantial investment in their development by both academic and industrial research groups, such that it may soon be feasible to produce a capsule endoscope with most of the functionality of current flexible endoscopes.\\ \\
Although robotic capsule endoscopy has high potential, it continues to face many challenges.  In particular, there is no broadly accepted method for generating a 3D map of the organ being investigated. This problem is made more severe by the fact that such a map may require a precise localization method for the endoscope, and such a method will itself require a map of the organ, a classic chicken-and-egg problem \citep{41}. The repetitive texture, lack of distinctive features, and specular reflections characteristic of the GI tract exacerbate this difficulty, and the non-rigid deformities introduced by peristaltic motions further complicate the reconstruction task \citep{42}. Finally, the small size of endoscope camera systems implies a number of limitations, such as restricted fields of view, low signal-to-noise ratio, and low frame rate, all of which degrade image quality \citep{43}. These issues, to name a few, make accurate and precise localization and reconstruction a difficult problem and can render navigation and control counterintuitive \citep{44}, \citep{46}.\\ \\
Despite these challenges, accurate and robust three-dimensional (3D) mapping of patient-specific anatomy remains a tantalizing goal. Such a map would provide doctors with a reliable measure of the size and location of a diseased area, thus allowing more intuitive and accurate diagnoses. In addition, should next-generation medical devices be actively controlled, a map would dramatically improve the doctor’s control in diagnostic, prognostic, and biopsy-like operations. As such, considerable energy has been devoted to adapting computer vision techniques to the problem of in-vivo 3D reconstruction of tissue surface geometry. \\ \\
Two primary approaches have been pursued as workarounds for the challenges mentioned previously. First, tomographic intra-operative imaging modalities, such as ultrasound (US), intra-operative computed tomography (CT), and interventional magnetic resonance imaging (iMRI) have been investigated for capturing detailed information of patient-specific tissue geometry \citep{45}. However, surgical and diagnostic operations pose significant technological challenges and costs for the use of such devices, due to the need to acquire a high signal-to-noise ratio (SNR) in real-time without impediment to the doctor.  Another proposal has been to equip endoscopes with alternative sensor systems in the hope of providing additional information; however, these alternative systems have other restrictions that limit their use within the body. \\ \\
This paper proposes a complete pipeline for 3D visual map reconstruction using only RGB camera images, with no additional sensor information. This pipeline is arranged in a modular form, and includes a preprocessing module for removal of specular reflections, vignetting and radial lens distortions, an image-stitching module for registration  of images, and a shape-from-shading (SfS) module for reconstruction of 3D structures. We provide  qualitative and quantitative analysis of pose estimation and 3D mapping accuracy using a real pig stomach, an esophagus gastro-duodenoscopy simulator, four different endoscopic camera models, an optical motion tracker and a 3D optical scanner. In sum, our method proposes a substantial contribution towards a more general, therapeutically relevant and extensive use of the information that capsule endoscopes may provide.

\section{LITERATURE SURVEY}
Several studies in literature have discussed 3D visual map reconstruction  for standard hand-held and passive capsule endoscopes \citep{2}, \citep{3},\citep{4}, \citep{5}, \citep{6}, \citep{8}, \citep{9}, \citep{10} etc.  These methods may be broken into four major classes; i.e 
\begin{itemize}
	\item stereoscopy
	\item shape-from-shading (SfS)
	\item structured light (SL)
	\item time-of-flight (ToF)
\end{itemize}
Structured light and time-of-flight methods require additional sensors, with a concordant increase in cost and space; as such, they are not covered in this paper. Stereo-based methods use the parallax observed when viewing a scene from two distinct viewpoints to obtain an estimate of the distance from observer to object under observation. Typically, such algorithms have four stages in computing the disparity map \citep{1}: cost computation, cost aggregation, disparity computation and optimization, and disparity refinement.\\ \\ With multiple algorithms reported per year, computational stereo depth perception has become a saturated field. The first work reporting stereoscopic depth reconstruction in endoscopic images was the work done by \citep{2}, which was implemented a dense computational stereo algorithm.  Later, Hager et al. developed a semi-global optimization \citep{3}, which was used to register the depth map acquired during surgery to pre-operative models \citep{4}.  Stoyanov et al. used local optimization to propagate disparity information around feature-matched seed points, and it has also been reported to perform well for endoscopic images.  This method was able to ignore highlights, occlusions or noise regions. Similar to stereo vision, another method that employs epipolar geometry and feature extraction is also proposed in \citep{30}. Similar to the stereo vision, this work flow starts with camera calibration and it mostly relies on SIFT extraction and feature description. Finally, the main algorithm calculates the 3D spatial point location using extrinsic parameters which is calculated from matched features in consecutive frames. Although this system exploits the advantage of sparse 3D reconstruction, the strong dependency on feature extraction causes performance related issues for endoscopic type of imaging. Despite the variety of algorithms and simplicity of implementation, computational stereo techniques bear several important flaws. To begin with, stereo reconstruction algorithms generally require two cameras, since the triangulation needs a known baseline between viewpoints.  Further, the accuracy of triangulation decreases with distance from the cameras due to the shrinkage of relative baseline between camera centers and reconstructed points (Micro-baseline problem). Most endoscopic capsule robots mount only one camera, and in those that mount more, the diameter of endoscope inherently bounds the baseline.  As such, stereo techniques have yet to find wide application in endoscopy.\\ \\
Due to the difficulty in obtaining stereo-compatible hardware, efforts have been made to adapt passive monocular three-dimensional reconstruction techniques to endoscopic images.
\begin{figure*}
	\includegraphics[width=0.90\textwidth]{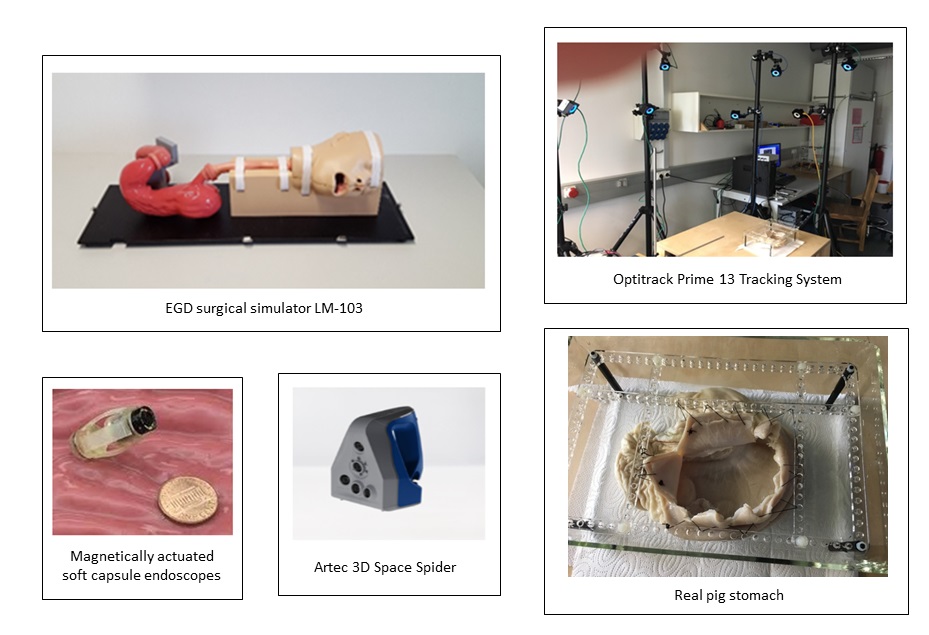}
	\caption{Schematics of the experimental setup for 3D visual map reconstruction of and a real pig stomach and an esophagus gastro-duodenoscopy simulator for surgical training, open surgical stomach model, 3D image scanner, endoscopic camera and active robotic capsule endoscope.}
	\label{fig:02}       
\end{figure*}
These techniques have been a focus of research in computer vision for decades, and have the distinct advantage of not requiring extra hardware equipment in addition to existing endoscopic devices.  Two main methods have emerged as useful in the field of endoscopic images: Shape-from-Motion (SfM) and Shape-from-Shading (SfS). SfS, which has been studied since the 1970s \citep{7} has demonstrated some suitability for endoscopic image reconstruction.  Its primary assumption is that there's a single lightsource on the scene, of which the intensity and pose relative to the camera are known. Both assumptions are mostly fulfilled in endoscopy \citep{8}, \citep{9}, \citep{10}. Further, the transfer function of the camera may be included in the algorithm to additionally refine estimates \citep{11}. Additional assumptions are that the object reflects light obeying Lambertian rules and that the object surface has a constant albedo. If these assumptions hold up to a degree and the equation parameters are known, SfS can use the brightness of a pixel to estimate the angle between camera’s depth axis and the shape normal at that pixel. This has been demonstrated to be effective in recovering details, although global shape recovery often has flaws.\\ \\
Both methods have been demonstrated to have flaws: SfS often fails in the presence of uncertain information, e.g. bleeding, reflections, noise artifacts and occlusions; feature tracking based SfM methods tend to fail in the presence of poorly textured areas and occlusions.\\ \\ Therefore, many state of art works are mainly based on the combination of these two techniques: In \citep{28}, a complete pipeline for 3D reconstruction of endoscopy imaging using SfS and SfM techniques is presented. In this work, the pipeline starts with basic preprocessing steps and focuses on 3D map reconstruction which is independent of light source position and illumination. Finally, the framework ends with frame-to-frame feature matching to solve the scaling issue of mono camera. This paper proposes interesting methods for the arduous task of reconstruction, however, more enhanced preprocessing and especially less dependency to feature extraction and matching is still needed. In recent work of \citep{39}, SfS and SfM are fused together to reach a better 3D map accuracy. With SFM, a sparse point cloud is obtained and dense version of this cloud is generated by the help of SFS. For better performance of SFS, they also propose refined reflectance model. One notable idea based on SfS and SfM fusion is proposed in \citep{31}. This methodology firstly reconstructs sparse 3D map points using SfM and iteratively refines the final reconstruction map using SfS. The approach does not directly attack the impediments caused by the ill posed illumination and specular reflectance, although, proposed geometric fusion tries to eliminates such defects. And the strong reliance on feature correspondence establishment remains yet unsolved. Attempts to solve the latter problem with template-matching techniques have had some success, but tend to be computationally very complex which makes it unsuitable for real-time performance. In \citep{32}, only SFS is used for reconstruction and 2D features are preferred for estimating transformation. Similarly, \citep{33} and \citep{34} combines SFM and SFS for 3D reconstruction without any preprocessing and with lambertian surface assumption. In \citep{35}, machine learning algorithms are applied for 3D reconstruction. Basically, training is completed with artificial dataset and real endoscopy images are used for test data. Another state of the art pipeline is proposed in \citep{36} which presents a work flow combining RGB camera and inertial measurement unit sensors (IMU). Besides improved results, this hardware makes the overall flow more complex and costly. Moreover, IMU sensors occupy extra place, they are not accurate enough and interfere with the magnetic actuation systems which makes them unsuitable for the next generation actively controllable endoscopic capsule robots. Main common issue remaining for 3D reconstruction of endoscopic-type datasets is the visual complexity of these images. These challenges which we mentioned in abstract and introduction cripples the performance of standard computer vision algorithms.  In particular, a proposed method must be robust to specular view-dependent highlights, noise, peristaltic movements, and focus-dependent changes in calibration parameters. Unfortunately, a quantitative measure of algorithm robustness has not been suggested in literature until today, despite its clear value towards evaluation of algorithmic dependability and precision. Moreover, all of the mentioned methods in that section were developed and evaluated on only one specific camera model, which  makes it impossible to justify the robustness of the framework in case of different camera choices with limited specifications such as lower resolution and image quality etc.\\ \\
Our paper proposes a full pipeline consisting of camera calibration, reflection detection and suppression, radial undistortion, de-vignetting, frame stitching, and SfS to reconstruct a 3D map of the organ under observation. Both synthetic and real stomachs are used for evaluations. Amongst other contributions, an extensive quantitative analysis has been proposed and enacted to demonstrate the influence of different pipeline modules on the accuracy and robustness of the reconstructed 3D map. To our knowledge, this is the first such comprehensive quantitative analysis to be enacted in endoscopic type of image processing.

\section{METHOD}
This section represents the proposed framework qualitatively and quantitatively in depth. Preprocessing steps, stitching module and SfS module will be discussed in detail.

\subsection{Preprocessing}
The proposed modular endoscopic 3D map reconstruction framework starts with a comprehensive preprocessing module which suppresses reflections caused by inner organ fluids, eliminates radial distortions and finally de-vignettes the images. Eliminating specular artifacts is a fundamental endoscopic image preprocessing step due to the accumulative errors caused by such distortions affecting the reconstructed final 3D map accuracy. For the reflection detection task, we propose an original method which identifies specular regions by combining shape and appearance information (see Fig. \ref{fig:04}). To extract the shape information of the reflection areas from gray-scale image, the gradient map of the input image is created and a morphological closing operation is applied on this map to fill the gaps inside reflection distorted areas. For the closing operation, we used OPEN CV function close(). In parallel to that shape-based approach, an appearance based method applies adaptive thresholding determined by the mean and standard deviation of the gray-scale image $I$ to identify the specular regions:
\begin{equation}
Mask_\textrm{$Illu$ = }\Bigg\{ 
\begin{tabular}{c c}
0 & ,$I < \mu_I+\sigma_I$\\
1 & ,otherwise \\
\end{tabular}  
\end{equation}
$\mu_I$, and $\sigma_I$ are the mean and standard deviation, respectively, of the image $I$. Combining both appearance based and shape based reflection detection parts using AND operation leads to a robust reflection detection performance. Once specular reflection pixels are detected, the inpainting method proposed by \citep{21} is applied to suppress the saturated pixels which replaces the specularity by an intensity value derived from a combination of neighboring pixel values
\begin{figure*}
	\includegraphics[width=0.90\textwidth]{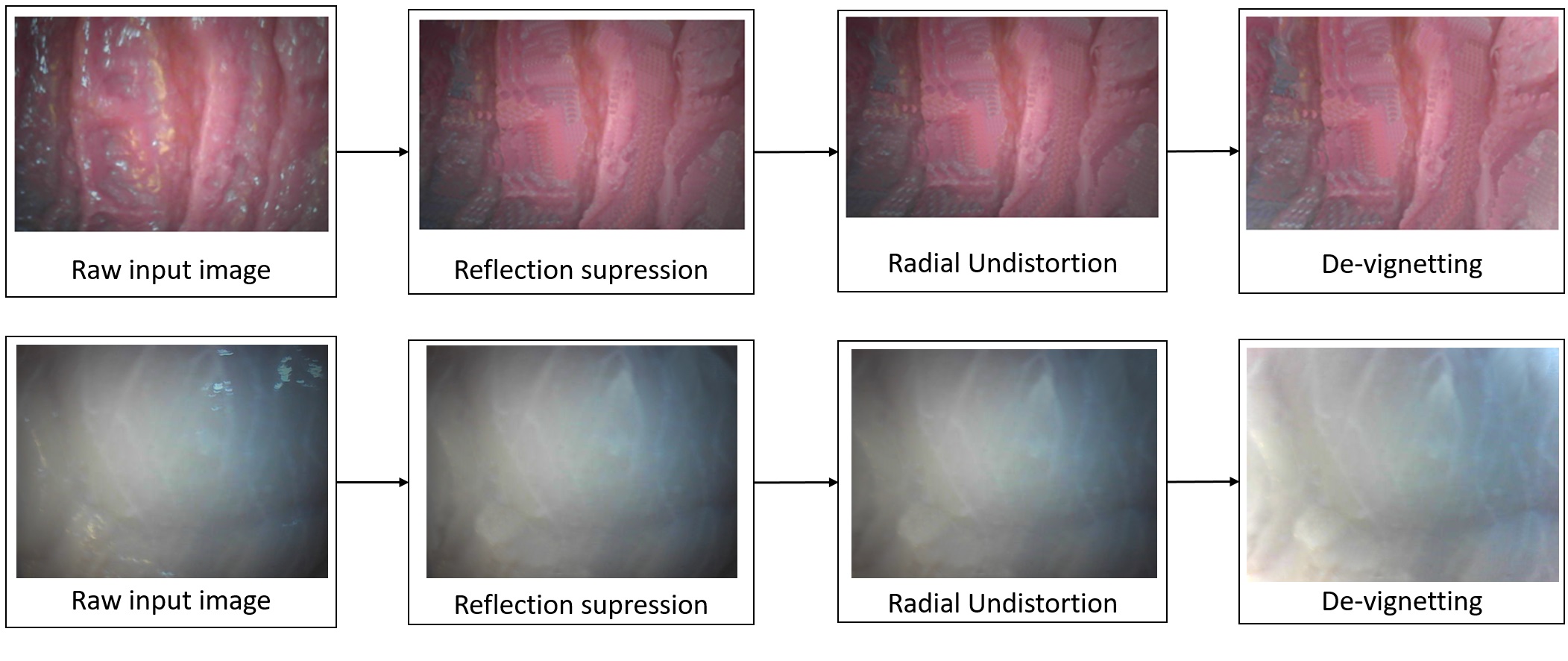}
	\caption{Preprocessing pipeline: Reflection removal, radial undistortion, de-vignetting}
	\label{fig:sfs}       
\end{figure*}
\begin{figure*}
	\includegraphics[width=0.90\textwidth]{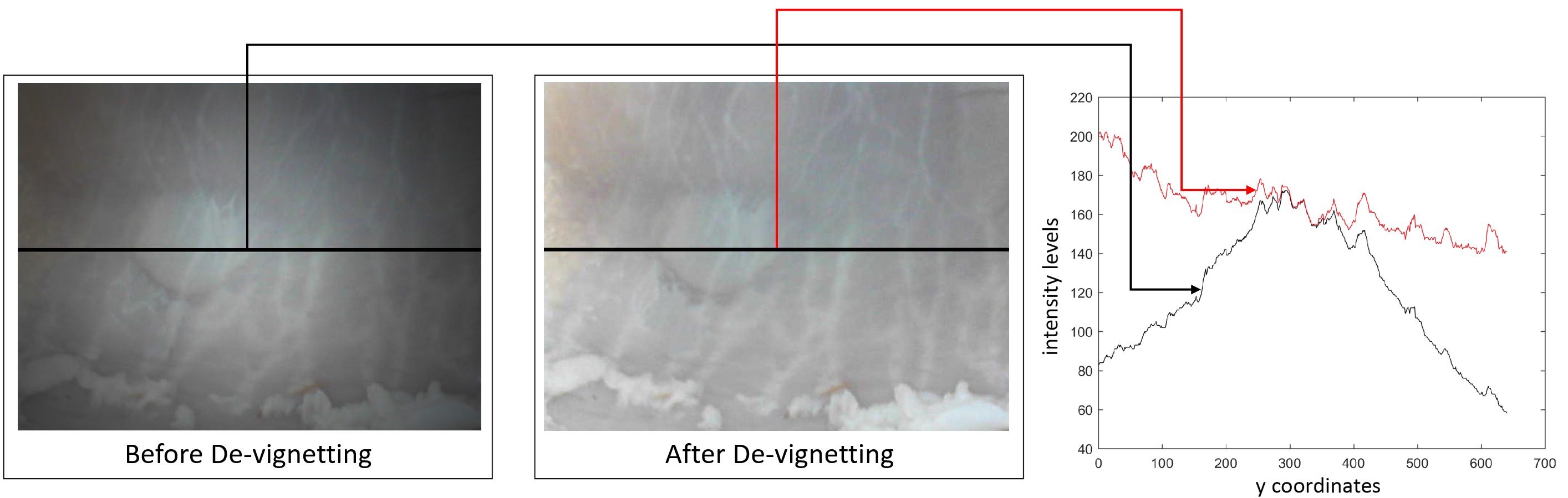}
	\caption{Demonstration of the de-vignetting process}
	\label{fig:vig}       
\end{figure*}
Distortion parameters obtained by the chessboard calibration method were used to remove radial lens distortions. The Brown-Conrady undistortion technique was applied to handle the radial distortions \citep{22},\citep{23}. Another common artifact of endoscopic type of images called vignetting which is referring to an inhomogeneous illumination distribution on the image corners with respect to image centers primarily caused by camera lens imperfections and light source limitations was handled by applying a radial gradient symmetry enforcement based method. Especially photometric pose estimation methods are very sensitive to vignetting artifacts, so a robust de-vignetting operation is required before proceeding to pose estimation steps. Our framework applies the vignetting correction approach proposed by \citep{15} which de-vignettes the image by enforcing the symmetry of the radial gradient from center to boundaries. An example of input image and vignetting-corrected output image may be seen in Fig. \ref{fig:sfs}.  The effect of de-vignetting is demonstrated in Fig. \ref{fig:vig}, where it is clearly observable that the intensity levels of de-vignetted image show a more homogenous pattern.

\subsection{Evaluation of Modern Feature Descriptors}
In that section, we evaluate the matching capability of feature point descriptors SURF, SIFT, HOG, ORB and two state-of-the art Optical Flow (OF) methods, Farneback and Lucas-Kanade regarding their matching accuracy on endoscopic type of images. We used OPENCV library for all of the implementations in that section. The re-projection error was used as the accuracy criteria for the performance evaluations \citep{16}. We evaluated the re-projection error over more than 500 endoscopic frame pairs carefully chosen from our real pig stomach dataset; results displayed in table \ref{tab:24} indicate that Farneback OF outperforms all other OF and feature based matching methods concerning the re-projection error minimization on endoscopic images. Moreover, one major issue of feature point based tracking approaches (SIFT, SURF and ORB) for endoscopic images was the insufficient amount of reliable feature points to execute a robust pose estimation. Poor textures, blurriness, sensor occlusions and noise artifacts are main factors which make feature point based tracking methods rather unsuitable for tracking tasks on endoscopic type of images. 
\begin{algorithm}
	\caption{Re-projection error calculation}
	\begin{algorithmic}[1]
		\STATE Extract and match feature points between two images using the selected feature descriptor/OF method (SIFT, SURF, ORB, Farneback and Lucas-Kanade Tracker).
		\STATE Perform homography estimation using RANSAC,  corresponding feature point pairs and intrinsic camera matrix
		\STATE Use the homography estimation to re-project the key point locations from the second image onto the first image.
		\STATE Calculate the re-projection error between the re-projected and initial key point locations.
	\end{algorithmic}
\end{algorithm}
\begin{table}
	\caption{Re-projection error of ORB, SURF, SIFT  Farneback and Lucas-Kanade Tracker for different endoscopic cameras.}
	\label{tab:24}       
	\newcolumntype{C}[1]{>{\centering\arraybackslash}p{#1}}
	%
	\begin{tabular}{C{2.5cm}C{1.5cm}C{1.5cm}C{1.7cm}C{1.5cm}C{0.5cm}}
		\hline\noalign{\smallskip}
		&Potensic&Misumi-I&Misumi-II&Awaiba\\
		\noalign{\smallskip}\noalign{\smallskip}
		Farneback  &0.11 &0.13 &0.12 &0.21&\\
		Lucas Kanade &0.13 &0.14 &0.14 &0.22&\\
		SIFT &0.16 &0.18 &0.17 &0.31&\\
		SURF &0.24 &0.26 &0.28 &0.45&\\
		ORB  &0.35 &0.37 &0.35 &0.55&\\
		\noalign{\smallskip}\hline\noalign{\smallskip}
	\end{tabular}
\end{table}

\subsection{Key frame selection}
Endoscopic videos generally contain thousands of highly overlapping frames (more than $\% 75$ overlap) due to incremental slow endoscopic capsule motion during organ exploration. A subset of the most relevant frames has to be chosen automatically which is called key frames. The minimum amount of key frames needed to recover the whole stomach surface with approx. $\%50$ overlapping area between frames is around 300 frames. Thus, at least every tenth frame could be selected as key frame. However, since the endoscopic capsule robot motion is not constant during organ exploration, it is not a good approach to simply choose every tenth frame in constant intervals. Our frame selection procedure is based on the optical flow vector interpretation between frame pairs. We sum up optical flow vector values of each pixel, and normalize the sum by the total pixel number. If the normalized sum is not exceeding a pre-defined threshold (30 pixels), it indicates that there is a high overlap between the corresponding frame pair (more than $\% 75$ overlap). In that case, the algorithm goes to the next frame and the loop starts again. The key frame selection procedure is represented by the algorithm 2:

\begin{algorithm}
	\caption{Frame selection algorithm}
	\begin{algorithmic}[1]
		\STATE Extract Farneback optical flow between reference key frame and candidate key frame
		\STATE Calculate the magnitude of the optical flow vector for every pixel.
		\STATE Sum up the magnitude values to find out the cumulative optical flow value.
		\STATE Divide the cumulative value by total pixel number for a normalization.
		\STATE If the normalized cumulative optical flow value is less than predefined threshold $\tau =20$ pixels, go to the next frame. Else identify the frame as a key frame and go tho the first step.
		\STATE If fifteen frames failed to fulfill the key frame conditions, and still $\tau =30$ pixels could not be exceeded, assign the frame with highest $\tau$ value among these fifteen frames as a  key frame and go to the next step.
	\end{algorithmic}
\end{algorithm}

\subsection{Frame stitching}

A state-of-the art stitching pipeline contains several stages:  Feature detection, which detects features within input images, feature matching, which matches features between input images, homography estimation, which estimates extrinsic camera parameters between pairs of matched images, bundle adjustment, which solves for all camera parameters jointly, image warping, which warps the images onto a compositing surface, gain compensation, which normalize the brightness and contrast of all images and finally blending, which blends pixels along the stitch seam to reduce the visibility of seams. Stitching algorithms fall broadly into two categories: direct alignment based and feature based methods. Direct alignment based methods attempts to match every pixel between the frame pair iteratively. This method has the benefit of using all the available data (which is a big advantage for low texture images such as endoscopic type of images), but it does require a good initialization and is very susceptible to varying brightness conditions. Feature based methods, on the other hand, first finds unique feature points such as corners and tries to match them. This method does not require an initialization, but the features found can be susceptible to illumination, zoom-in and out and rotation changes. Our frame stitching technique does not rely on feature points and works in a coarse-to-fine fashion combining Farneback OF based coarse alignment with patch-wise intensity based fine alignment. Farneback OF combined with RANSAC finds the homography which is the initial camera motion estimation, whereas the SSD-based energy minimization applied on circular regions-of-interest with a radius of 15 pixels around each inlier point after RANSAC refines the estimation. For the RANSAC method, we use the homography estimation between the frame pair using a set of corresponding points derived from Farneback OF. The fine image registration method estimates the parameters of affine transformation by minimizing an intensity difference based energy cost function. The affine transformation maps an image $I_1$ onto the reference image $I_2$, where $x'$ , $y'$ represent the transformed and $x$, $y$ the original pixel coordinates, and $a_1$ , $a_2$ , $a_3$ , $a_4$ , $t_x$ , $t_y$  the parameters of affine transformation matrix $A$, respectively. We define a cost function measuring the pixel intensity similarity between the image pair (see Eq. \ref{eq:cost}), which is supposed to be minimized by the corresponding affine transformation parameters. 
\begin{equation}
\begin{pmatrix}
x_2\\ 
y_2\\ 
1
\end{pmatrix}
=\begin{pmatrix}
a_1 &a_2 &t_x \\ 
a_3& a_4 &t_y \\ 
0&0&1 
\end{pmatrix}\cdot \begin{pmatrix}
x_1\\ 
y_1\\
1 
\end{pmatrix}
\end{equation}
Since the cost function has to ignore the pixels lying outside the circular patches defined around inlier points, a weighting function $w(x, y)$ is defined:
\begin{equation}
\omega (x,y)=\left\{\begin{matrix}
0, if   (x-x_c)^{2}+(y-y_c)^{2}\geq r^{2}\\ 
1, if   (x-x_c)^{2}+(y-y_c)^{2}<  r^{2}
\end{matrix}\right.
\end{equation}
where $x_c$  and $y_c$ are the coordinates of inlier point and $r$ the radius of circular image region around this inlier point center. The resulting cost function has a bias toward smaller overlap solutions, thus a normalization of it by the overlap area is necessary, resulting in the mean squared pixel error (MSE):
\begin{equation}
\label{eq:cost}
e_{MSE}(A)=\frac{\sum_i \omega (x_i,y_i)\omega (x_i',y_i')(I_2(x_i',y_i')-I_1(x_i,y_i))^{2}}{\sum_i{\omega (x_i,y_i)\omega (x_i',y_i')}}
\end{equation}
The affine transformation matrix $A$ is iteratively determined by the image transformation that minimizes $e_{MSE}$ using Gaussian-Newton Optimization. CUDA library was utilized to achieve better performance and reduce timing through parallelism. Now that our cost function is defined, an efficient search strategy for the global minimum has to be determined to avoid the Gaussian-Newton Optimization resulting in false convergences. From the six affine transformation parameters, the most significant ones are the two translations in x-and y-direction, tx and ty, because endoscopic scanning procedure contains dominantly translational motions. Thus, the first step of our algorithm gives a rough estimate of $t_x$ , $t_y$ using the OF vectors $u$ and $v$ acquired by Farneback OF.  Once a rough initialization is done, the second step is the estimation of all affine transformation parameters, using the iterative Gaussian-Newton Optimization to minimize $e_{MSE}$. 

\begin{figure*}
	\centering
	\includegraphics[width= 1\textwidth]{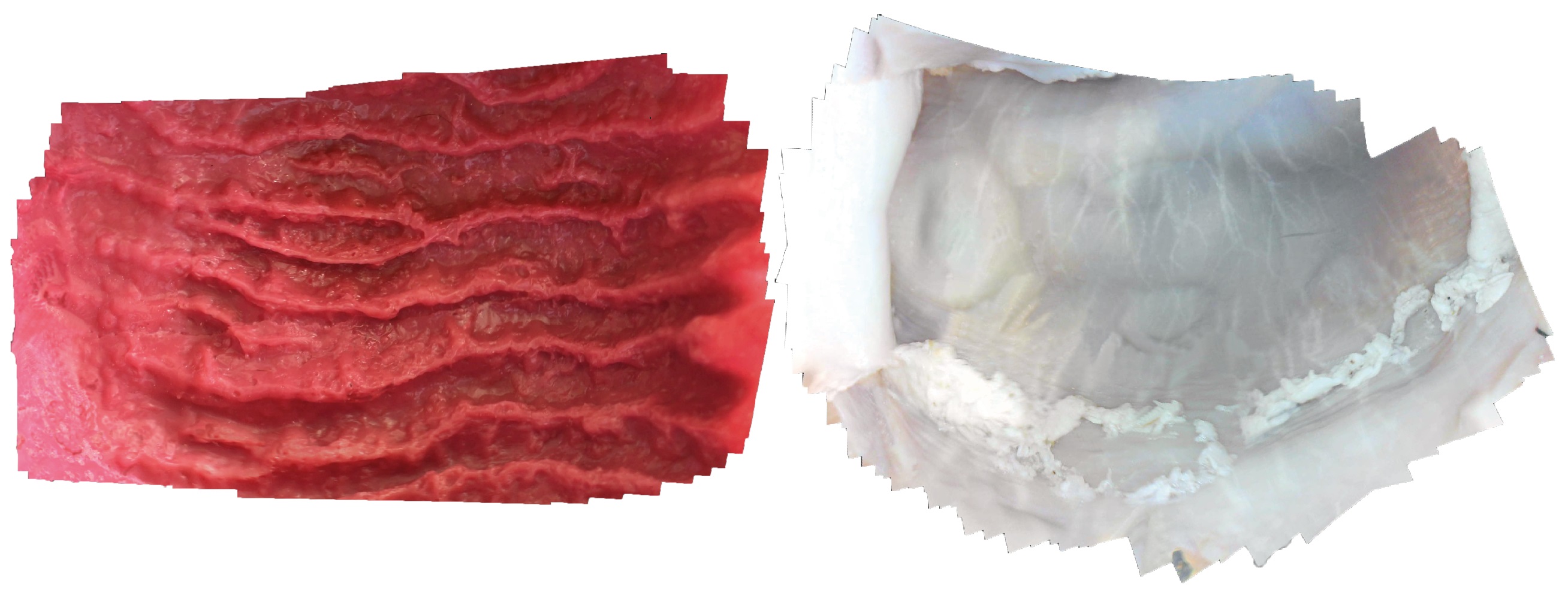}
	
	\caption{ Demonstration of the frame stitching process for non-rigid esophagus gastro duodenoscopy simulator (left) and real pig stomach (right).}
	\label{fig:09}       
\end{figure*}

\begin{figure*}
	\includegraphics[width= 1\textwidth]{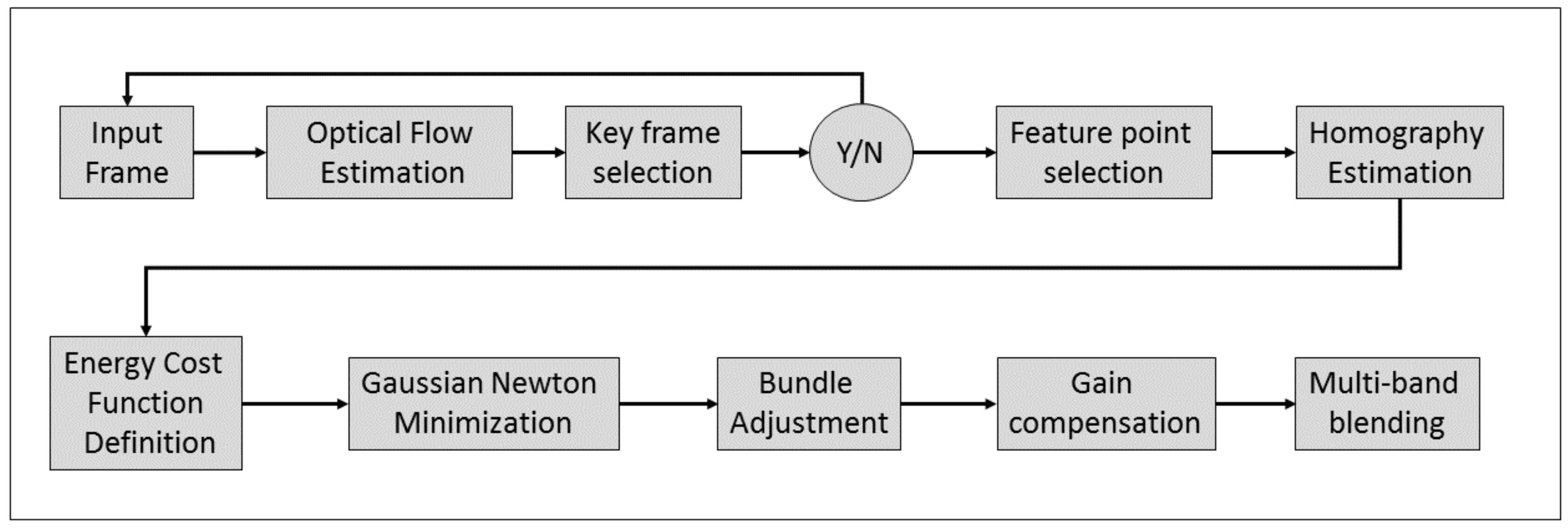}
	\caption{Image stitching flowchart.}
	\label{fig:0ff9}       
\end{figure*} 
The termination criteria of Gaussian Newton optimization was defined by a threshold $\tau=e^{-9}$, whereas the optimization stops when the $e_{MSE}$ drops below that threshold or maximum number of iterations are already reached (100 iterations). Once the optimization has converged and the affine transformation parameters are estimated, bundle adjustment will be executed to solve the registration problem for all the camera parameters jointly and to correct potential accumulative errors.  For the mosaic image representation, all key frames $I_i$ are transformed into the coordinate system of the anchor key frame $I_A$. In areas where several key frames overlap, corresponding image pixels often do not have the same intensity and color due to inhomogeneous illumination, intensity variations due to preprocessing steps, focusing and de-focusing etc. An advanced image blending algorithm has to be applied to overcome these issues and create a high quality final mosaic image. We used multi-band blending method proposed by \citep{37} which preserves high frequency information of endoscopic images and suppresses low frequency variations caused by irregular illumination. The blending algorithm overview is to see in Fig. \ref{fig:0f9}. For details of this multi-band blending technique, the reader is referred to the original work of \citep{37}. The following steps in algorithm 3 were applied for the mosaicking process: 
\begin{figure*}
	\includegraphics[width= 1\textwidth]{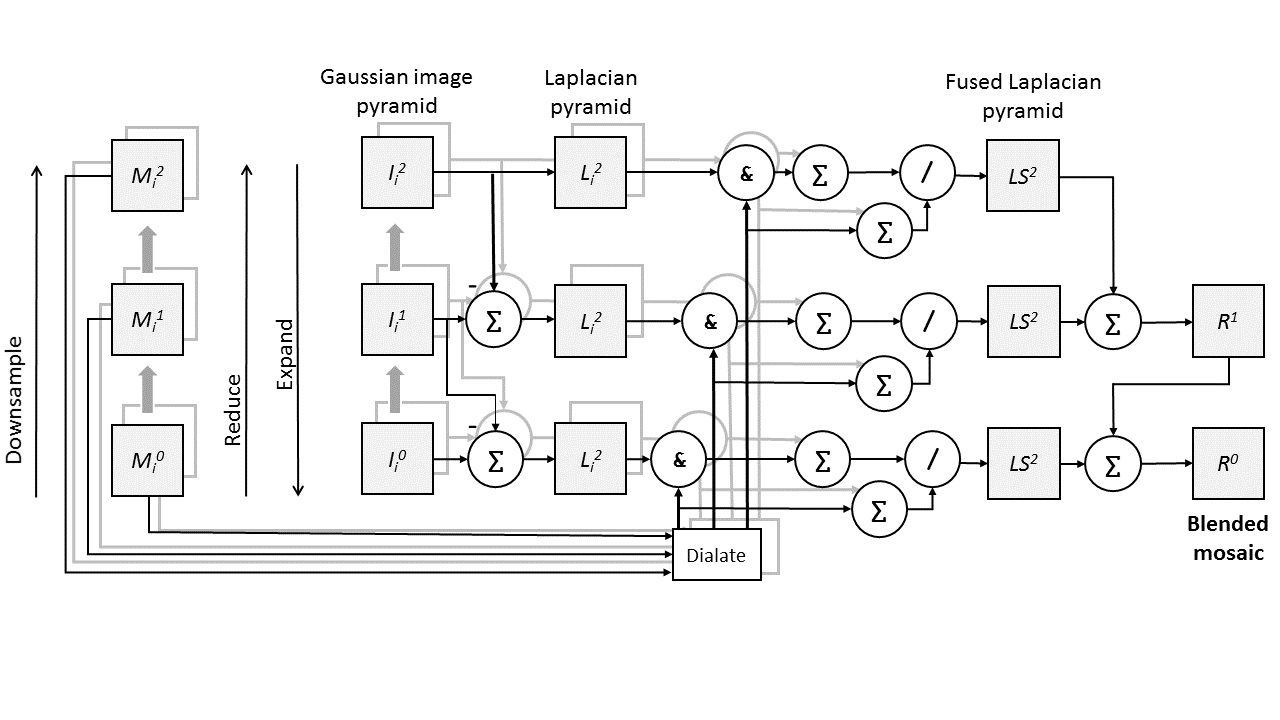}
	\caption{Multi-band blending flowchart.}
	\label{fig:0f9}       
\end{figure*}

\begin{algorithm}
	\caption{Coarse-to-fine image mosaicking}
	\begin{algorithmic}[1]
		\STATE Execute key frame selection module to identify a key frame.
		\STATE Match pixels using optical flow estimation.
		\STATE Use RANSAC and matched points to estimate the homography (coarse pose estimation)
		\STATE Define circular regions around each inlier point.
		\STATE Calculate the intensity difference based energy cost function.
		\STATE Execute iterative Gaussian Newton optimization to minimize the energy cost function. Use optical flow vector as an initial point for the GN optimization. 
		\STATE Perform bundle adjustment to globally optimize all of the camera poses jointly. (GPU based multi-core bundle adjustment of \citep{38} was applied in our framework.)
		\STATE Perform frame warping 
		\STATE Perform gain compensation (\citep{40})
		\STATE Perform multi-band blending. (Multi-resolution spline based blending technique of \citep{37} was employed in our framework.)
	\end{algorithmic}
\end{algorithm}

\subsection{Depth image creation}
Once the final mosaic image is obtained, the next module creates its depth image using SfS
technique of Tsai and Shah [\citep{17}]. Tsai-Shah SfS method is based on the following assumptions:
\begin{itemize}
	\item	The object surface is lambertian
	\item	The light comes from a single point light source
	\item	The surface has no self-shaded areas. 
\end{itemize}
This first assumption is not obeyed by raw endoscopic images due to the specular reflections inside the organs. We addressed this problem through the reflection suppression technique previously described. Subsequently, the above assumptions allow the image intensities to be modeled by
\begin{equation}
I(x,y)=\rho(x,y,z)\cdot cos\Theta_i
\end{equation}
where $\textit{I}$ is the intensity value, $p$ is the albedo (reflecting power of surface), and theta is the angle between surface normal $\textit{N}$ and light source direction $\textit{S}$. With this equation, the gray values of an image $I$ are related only to albedo and angle theta.  Using these assumptions, the above equation can be rewritten as follows:
\begin{equation}
I(x,y)=\rho\cdot N.S
\end{equation}
where (.) is the dot product, $N$ is the unit normal vector of the surface, and $S$ is the incidence direction of the source light.  These may be expressed respectively as
\begin{equation}
N=\frac{(-p(x,y),-q(x,y),1)}{(p^2+q^2+1)^\textrm{(1/2)}}
\end{equation}
\begin{equation}
S=(cos\tau\cdot sin\sigma,sin\tau\cdot sin\sigma,cos\sigma  )
\end{equation}
where ($\tau$) and ($\sigma$) are the slant and tilt angles, respectively, and $p$ and $q$ are the $x$ and $y$ gradients of the surface $Z$:
\begin{equation}
\label{eq:12}
p(x,y) = \frac{\partial Z(x,y)}{\partial x}
\end{equation}
\begin{equation}
\label{eq:13}
q(x,y) = \frac{\partial Z(x,y)}{\partial y}
\end{equation}
The final function then takes the form
\begin{multline}
\label{eq:14}
I(x,y) = \rho \cdot \frac{(cos\sigma+p(x,y)\cdot cos\tau\cdot sin
	\sigma+q(x,y)\cdot sin\tau\cdot sin\sigma  )}{((p(x,y))^2+(q(x,y))^2+1)^\textrm{(1/2)}} 
\\= R(p_\textrm{x,y}, q_\textrm{x,y})
\end{multline}
Solving this equation for $p$ and $q$ essentially corresponds to the general problem of SfS. The approximations and solutions for $p$ and $q$ give the reconstructed surface map $Z$. The necessary parameters are tilt, slant and albedo, and can be estimated as proposed in \citep{16}. The unknown parameters of the 3D reconstruction are the horizontal and vertical gradients of the surface $Z$, $p$ and $q$. With discrete approximations, they can be written as follows:
\begin{equation}
\label{eq:15}
p(x,y)=Z(x,y)-Z(x-1,y)
\end{equation}
\begin{equation}
\label{eq:16}
q(x,y)=Z(x,y)-Z(x,y-1)  
\end{equation}
where $Z(x,y)$ is the depth value of each pixel.  From these approximations, the reflectance function R($p_\textrm{x,y}, q_\textrm{x,y}$) can be expressed as 
\begin{equation}
\label{eq:17}
R(Z(x,y)-Z(x-1,y),Z(x,y)-Z(x,y-1))         
\end{equation}
Using equations \ref{eq:15}, \ref{eq:16}, and \ref{eq:17}, the reflectance equation may also be written as
\begin{multline}
\label{eq:18}
f(Z(x,y),Z(x,y-1),Z(x-1,y),I(x,y))\\
= I(x,y)- R(Z(x,y)-Z(x-1,y),Z(x,y)-Z(x,y-1))\\
=0
\end{multline}
Tsai and Shah proposes a linear approximation using a first-order Taylor series expansion for function $f$ and for depth map $Z^\textrm{n-1}$, where$ Z^\textrm{n-1}$ is the recovered depth map after $n-1$ iterations. The final equation is
\begin{equation}
Z^n  (x,y)=Z^\textrm{(n-1)} (x,y)  -\frac{f(Z^\textrm{(n-1)} (x,y))}{\frac{\mathrm{d}(f(Z^\textrm{(n-1)} (x,y))}{\mathrm{d}(Z(x,y))}}
\end{equation}
where $f$ is a predefined function, constrained by
\begin{equation}
\frac{\mathrm{d}f(Z^\textrm{(n-1)}(x,y))}{\mathrm{d}Z(x,y)}√(1+i_x^2+i_y^2 ))
\end{equation}
and
\begin{equation}
i_x = cos\tau\cdot\frac{sin\sigma}{cos\sigma}
\end{equation}
\begin{equation}
i_y = sin\tau\cdot\frac{sin\sigma}{cos\sigma}
\end{equation}
The $n^{th}$ depth map $Z^n$ is calculated by using the estimated slant, tilt, and albedo values. Resulting sample images for reflection removal and SfS are to see in Fig. \ref{fig:sfs}

\section{Evaluation}
We evaluate the performance of our system both quantitatively and qualitatively in terms of trajectory estimation and surface reconstruction. We also report the computational complexity of the proposed framework.

\subsection{Dataset}
We created our dataset on a real pig stomach and a non-rigid open GI tract model EGD (esophagus gastro duodenoscopy) surgical simulator LM-103. We used the EGD surgical simulator for quantitative analyses, and the real pig stomach for qualitative evaluations. Synthetic stomach fluid was applied to the surface of the EGD simulator to imitate the mucosa layer of the inner tissue. A robust endoscopic localization and mapping framework should preserve its performance and functionality in case of varying camera specifications. To ensure that our algorithm is not tuned to a specific camera model, which is a common problem we observed on the proposed methods in literature, four different commercially available endoscopic cameras were employed for the video capture.  With that aim, we carefully selected four commercially available endoscopic cameras varying in their specifications (resolution, pixel size, image quality, blurriness etc). A total of 17010 endoscopic frames were acquired by these four camera models which were mounted on our robotic magnetically actuated soft capsule endoscope prototype (MASCE). The first sub-dataset, consisting of 4230 frames, was acquired with an Awaiba NanEye camera (see Table \ref{tab:1aa}). The second sub-dataset, consisting of 4340 frames, was acquired by the Misumi V3506-2ES endoscopic camera with the specification shown in Table \ref{tab:1ab}. The third sub-dataset of 4320 frames, was obtained by the Misumi V5506-2ES endoscopic camera with the specification shown in Table \ref{tab:1ac}. Finally, the fourth sub-dataset of 4120 frames, was obtained by the Potensic mini camera with the specification shown in Table \ref{tab:1acd}. We scanned the open stomach simulator using the 3D Artec Space Spider image scanner and used this 3D scan as the ground truth for the 3D map reconstruction framework (see Fig. \ref{fig:02}). Even though our focus and ultimate goal is an accurate and therapeutically relevant 3D map reconstruction, we also evaluated the pose estimation accuracy of the proposed framework quantitatively since a precise pose estimation is a pre-requisite for an accurate 3D mapping. Thus, Optitrack motion tracking system consisting of eight Prime-13 cameras and a tracking software were utilized to obtain 6-DoF localization ground truth data of the endoscopic capsule motion in a sub-millimeter precision (see Fig. \ref{fig:02}). 

\begin{figure}
	\centering
	\includegraphics[width=0.90\textwidth]{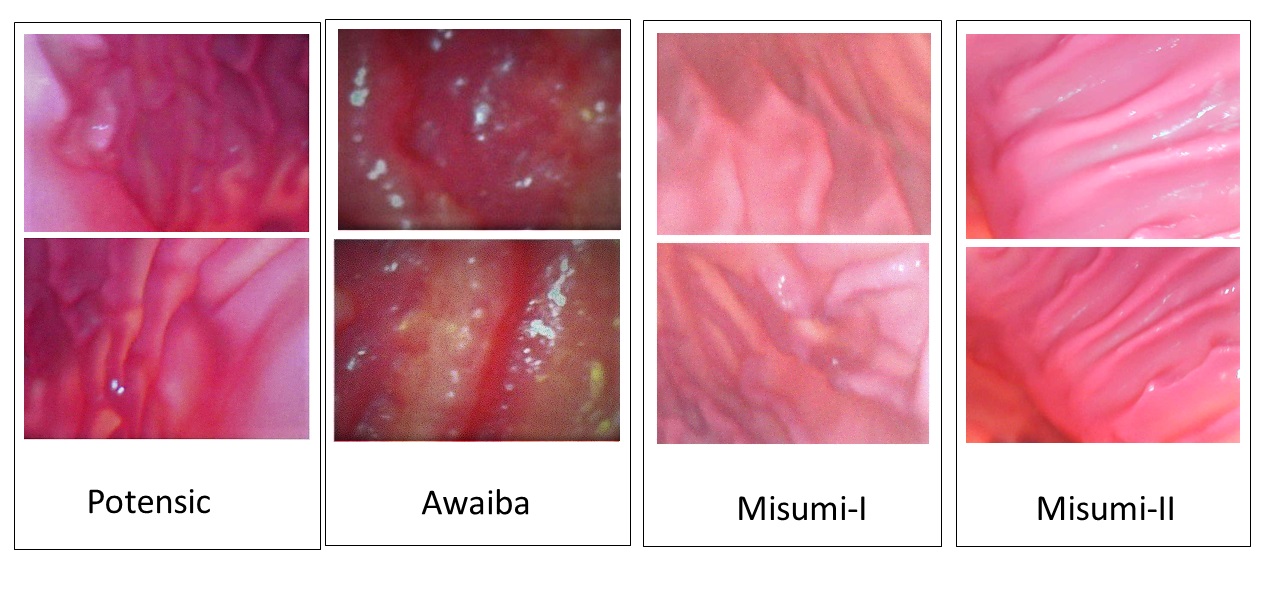}
	\caption{Non-rigid esophagus gastro duodenoscopy simulator dataset overview for different endoscopic cameras.}
	\label{fig:01}       
\end{figure}

\begin{figure}
	\centering
	\includegraphics[width=0.90\textwidth]{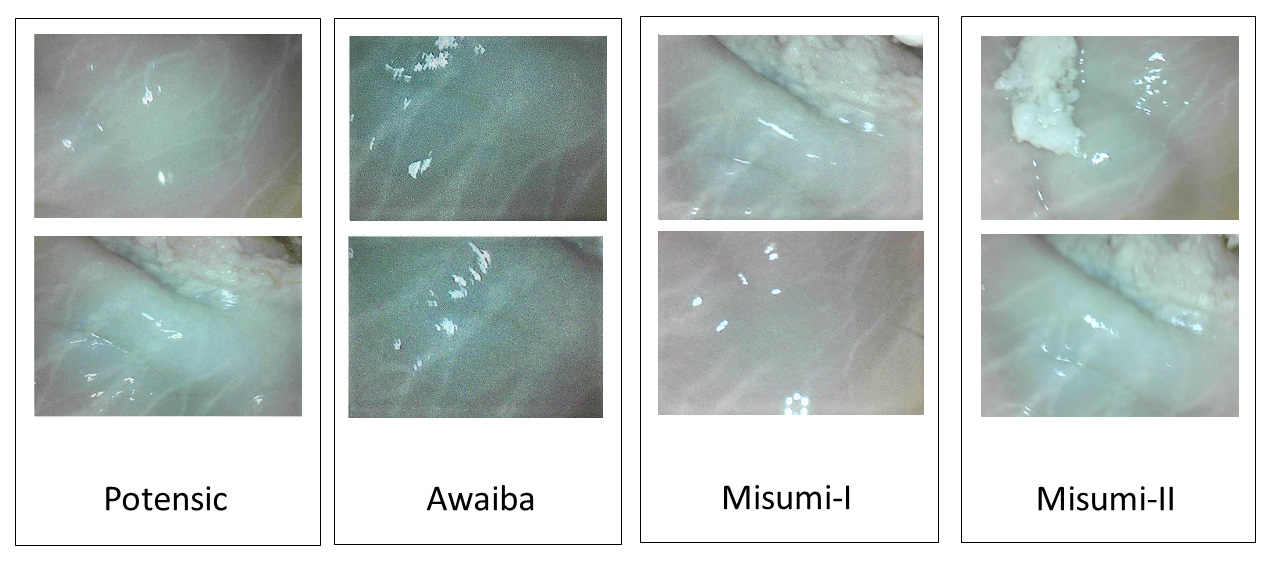}
	\caption{Real pig stomach dataset overview for different endoscopic cameras.}
	\label{fig:pig}       
\end{figure}

\begin{table}
	\centering
	\caption{Awaiba Naneye Monocular Endoscopic Camera }
	\label{tab:1aa}       
	%
	\begin{tabular}{p{2cm}p{2.4cm}p{2cm}p{4.9cm}}
		\hline\noalign{\smallskip}
		\\
		Resolution & 250 x 250 pixel& \\
		Diameter & 2.2mm \\
		Pixel size & 3 x 3 $\mu m^2$\\
		Pixel depth & 10 bit \\
		Frame rate& 44 fps\\
		\noalign{\smallskip}\hline\noalign{\smallskip}
	\end{tabular}
\end{table}

\begin{table}
	\centering
	\caption{Misumi-V3506-2ES monocular camera }
	\label{tab:1ab}       
	%
	\begin{tabular}{p{2cm}p{2.4cm}p{2cm}p{4.9cm}}
		\hline\noalign{\smallskip}
		\\
		Resolution & 400 x 400 pixel& \\
		Diameter & 8.2mm \\
		Pixel size & 5.55 x 5.55 $\mu m^2$\\
		Pixel depth & 10 bit \\
		Frame rate& 30 fps\\
		\noalign{\smallskip}\hline\noalign{\smallskip}
	\end{tabular}
\end{table}

\begin{table}
	\centering
	
	\caption{Misumi-V5506-2ES monocular camera }
	\label{tab:1ac}       
	%
	\begin{tabular}{p{2cm}p{2.4cm}p{2cm}p{4.9cm}}
		\hline\noalign{\smallskip}
		\\
		Resolution & 640 x 480 pixel& \\
		Diameter & 8.6 mm \\
		Pixel size & 6.0 x 6.0 $\mu m^2$\\
		Pixel depth & 10 bit \\
		Frame rate& 30 fps\\
		\noalign{\smallskip}\hline\noalign{\smallskip}
	\end{tabular}
\end{table}
\begin{table}
	\centering
	\caption{Potensic Mini Monocular Camera }
	\label{tab:1acd}       
	%
	\begin{tabular}{p{2cm}p{2.4cm}p{2cm}p{4.9cm}}
		\hline\noalign{\smallskip}
		\\
		Resolution & 1280 x 720 pixel& \\
		Diameter & 8.8 mm \\
		Pixel size & 10.0 x 10.0 $\mu m^2$\\
		Pixel depth & 10 bit \\
		Frame rate& 30 fps\\
		\noalign{\smallskip}\hline\noalign{\smallskip}
	\end{tabular}
\end{table}

\subsection{Trajectory Estimation}
\begin{figure*}
	\includegraphics[width= 1\textwidth]{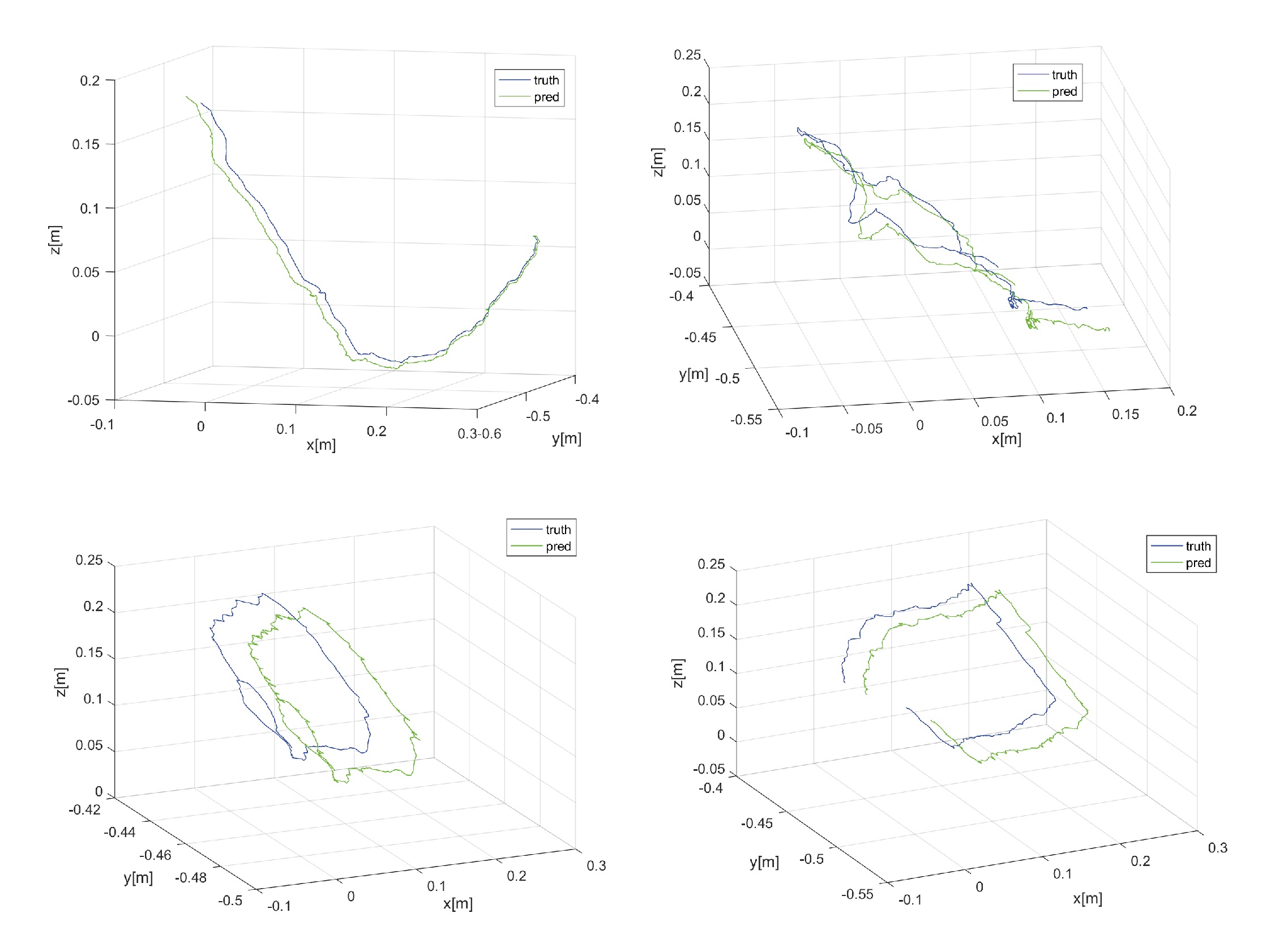}
	\caption{Qualitative pose estimation results.}
	\label{fig:0fg9}       
\end{figure*}

To evaluate the pose estimation performance, we test our system on different trajectories of various difficulty levels. The absolute trajectory (ATE) root-mean-square error metric (RMSE) is used for quantitative pose accuracy evaluations, which measures the root-mean-square of Euclidean distances between  estimated endoscopic capsule robot poses and the ground truth poses delivered by the motion capture system. Table \ref{tab:1a} demonstrates the results of trajectory estimation for six different trajectories. Trajectory 1 is an uncomplicated path with very slow incremental translations and rotations. Trajectory 2 follows a comprehensive scan of the stomach with many local loop closures. Trajectory 3 contains an extensive scan of the stomach with more complicated local loop closures. Trajectory 4 consists of more challenge motions including fast rotational and translational frame-to-frame motions. Trajectory 5 is the same of trajectory 4 but included synthetic noise to see the robustness of system against noise effects. Before capturing trajectory 6, we added more synthetic stomach oil into the simulator tissue to have heavier reflection conditions. Similar to the trajectory 5, trajectory 6 consists of very loopy and complex motions.  As seen in table \ref{tab:1a}, the system performs very robust and accurate in terms of trajectory tracking in all of the challenge datasets. Tracking accuracy is only decreased for very fast frame-to-frame movements, motion blur, noise or heavy spectral reflections occurring frequently in last trajectories especially. Qualitative pose estimation results shown in Fig. \ref{fig:0fg9} indicate that our coarse-to-fine pose estimation approach tracks the robot trajectory very accurately for different type of motions. 
\begin{table}
	
	\caption{Comparison of ATE RMSE for different trajectories and cameras in cm}
	\label{tab:1a}       
	\newcolumntype{C}[1]{>{\centering\arraybackslash}p{#1}}
	%
	\begin{tabular}{C{1.5cm}C{1.9cm}C{1.4cm}C{1.5cm}C{1.7cm}C{1.7cm}C{0.1cm}}
		\hline\noalign{\smallskip}
		& Length & Potensic & Misumi-I& Misumi-II & Awaiba\\
		\noalign{\smallskip}\noalign{\smallskip}
		Traj 1 & 123.5 & 4.10 & 4.23 &4.17 &6.93 &\\
		Traj 2 & 132.4 & 4.14 & 4.45 &4.32 &7.12 &\\
		Traj 3 & 124.6 & 5.23 & 5.54 &5.43 &7.42 &\\
		Traj 4 & 128.2 & 5.53 & 5.67 &5.47 &7.51 & \\
		Traj 5 & 128.2 & 6.32 & 5.45 &5.32  &8.32 &\\
		Traj 6 & 123.1 & 7.73 & 6.72 &6.51 &8.73 & \\
		\noalign{\smallskip}\hline\noalign{\smallskip}
	\end{tabular}
\end{table}
RMSE results for pose estimation before and after application of reflection suppression, de-vignetting, and radial un-distortion were evaluated and compared to quantitatively analyze their effects in terms of pose estimation accuracy. Results shown in table \ref{tab:1ad} for Misumi camera-II indicate that reflection suppression unexpectedly leads to a decrease of pose estimation performance. The reason for such a decrease after reflection suppression is related to the fact that such saturated peak values contain pose information which will be cut-off during reflection suppression. Since such saturated pixels do not change in their appearance between small base-line frame pairs drastically (very few alteration of the illumination incidence angle on the surface), existence of these reflection pixels might have resulted in a slightly better pose alignment. On the other hand, radial un-distortion and de-vignetting operations both increase pose estimation accuracy of the framework as expected.

\begin{table}
	
	\caption{Comparison of ATE RMSE (in cm) for MISUMI-II camera and different combinations of preprocessing operations, NPR: No preprocessing applied,  RS:Reflection suppression applied, RUD: Radial un-distortion applied, DV: De-Vignetting applied}
	\label{tab:1ad}       
	\newcolumntype{C}[1]{>{\centering\arraybackslash}p{#1}}
	%
	\begin{tabular}{C{2cm}C{2cm}C{2cm}C{2cm}C{2cm}C{0.1cm}}
		\hline\noalign{\smallskip}
		&RS&NRS&RS+RUD&RS+RUD+DV\\
		\noalign{\smallskip}\noalign{\smallskip}
		Traj 1  & 5.45& 4.12 &4.01 &4.03&\\
		Traj 2  & 6.44& 4.23 &4.07 &4.04&\\
		Traj 3  & 6.57& 5.13 &4.97 &4.98&\\
		Traj 4  & 7.55& 5.34 &5.16 &5.08&\\
		Traj 5  & 8.43& 5.43 &5.14 &5.02&\\
		Traj 6  & 8.69& 5.64 &5.25 &5.12&\\
		\noalign{\smallskip}\hline\noalign{\smallskip}
	\end{tabular}
\end{table}

\subsection{Surface Reconstruction}
\begin{figure*}
	\begin{center}
		\includegraphics[width=0.95\textwidth]{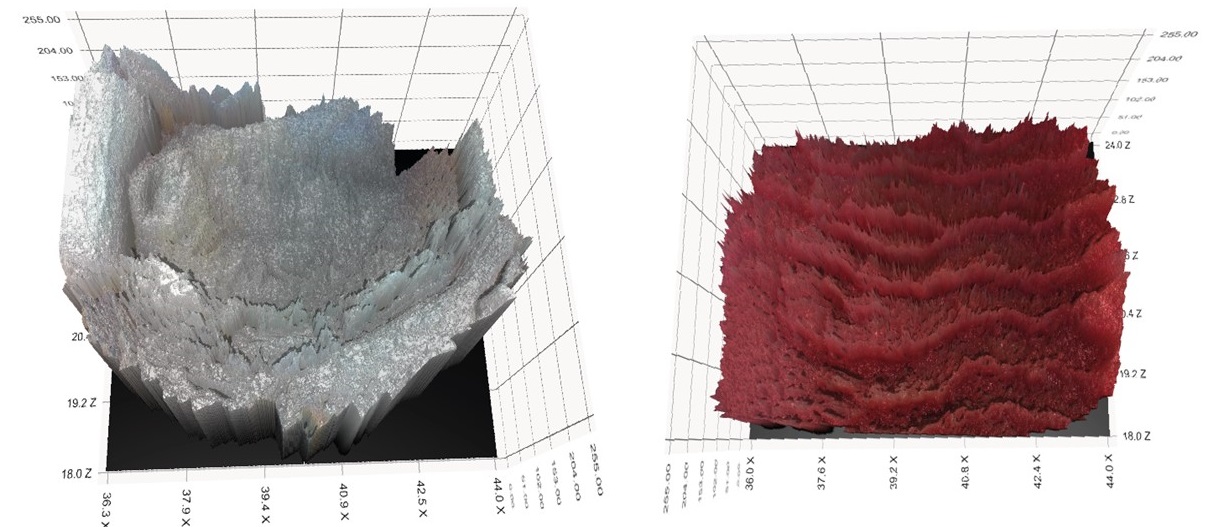}
		\caption{Qualitative 3D reconstructed map results for different cameras (Real pig stomach (left), synthetic human stomach (right)}
		\label{fig:map2}       
	\end{center}
\end{figure*}

We evaluated the surface reconstruction accuracy of our system on the same dataset we applied for the trajectory estimation framework as well. We scanned the open non-rigid EGD (Esophago-gastroduodenoscopy) simulator to obtain the ground truth 3D data using a highly accurate commercial 3D scanner (Artec 3D Space Spider). Final 3D map of the stomach model obtained by the proposed framework and the ground truth scan were aligned using iterative closest point algorithm (ICP). The absolute depth (ADE) RMSE was used to evaluate the performance of the map reconstruction approach, which measured the root-mean-square of Euclidean distances between estimated depth values and the corresponding ground truth depth values. A lowest RMSE of $2.14cm$ (see Table \ref{tab:2}) prove that our system can achieve very high map accuracies. Even in more challenge trajectories such as trajectory $3$, our system is still capable of providing an acceptable 3D map of the explored inner organ tissue. 3D reconstructed maps of real pig stomach and syntetic human stomach are represented in Fig. \ref{fig:map2} for visual reference.
\begin{table}
	\caption{Comparison of surface reconstruction accuracy (in cm) results on the evaluated datasets. Quantities shown are the mean distances from each point to the nearest surface in the ground truth 3D model in cm}
	\label{tab:2}       
	\newcolumntype{C}[1]{>{\centering\arraybackslash}p{#1}}
	%
	\begin{tabular}{C{1.5cm}C{1.5cm}C{1.5cm}C{1.5cm}C{1.5cm}C{1.5cm}C{0.5cm}}
		\hline\noalign{\smallskip}
		& Depth &Potensic&Misumi-I&Misumi-II&Awaiba\\
		\noalign{\smallskip}\noalign{\smallskip}
		Traj 1 &  63.42 &2.82 &2.32 &2.14 &3.42&\\
		Traj 2 &  63.45 &2.56 &2.45 &2.16 &4.14&\\
		Traj 3 &  63.41 &3.16 &2.76 &2.45 &4.45&\\
		\noalign{\smallskip}\hline\noalign{\smallskip}
	\end{tabular}
\end{table}
to indicate the contributions of each preprocessing module on the map reconstruction accuracy, map accuracy was tested after applying each preprocessing step and reconstructing the map. As shown in table \ref{tab:1af}, each preprocessing operation has a certain influence on the RMSE results. One important observation is that even though pose accuracy increases with existence of reflection points, these saturated pixels have negative influence on the map accuracy, as expected. Therefore, disabling reflection suppression during pose estimation and enabling it for map reconstruction is the best option to follow. 
\begin{table}
	\caption{Comparison of ATE RMSE (in cm) for different trajectories and different combinations of preprocessing operations on the evaluated dataset, NPR: No Prepocessing applied, RSPM: Reflection suppression applied for both pose estimation and map reconstruction, RSM: Reflection suppression applied only for map reconstruction, RUD: Radial undistortion applied, DV: De-Vignetting applied, MISUMI-II camera was used.  Quantities shown are the mean distances from each point to the nearest surface in the ground truth 3D model in cm}
	\label{tab:1af}       
	\newcolumntype{C}[1]{>{\centering\arraybackslash}p{#1}}
	%
	\begin{tabular}{C{1.2cm}C{1.2cm}C{1.2cm}C{1.8cm}C{2cm}C{2cm}C{0.5cm}}
		\hline\noalign{\smallskip}
		&NPR&RSM&RSPM&RSPM+RUD&RSPM+RUD+DV\\
		\noalign{\smallskip}\noalign{\smallskip}
		Traj 1  &5.45& 3.65&3.42 &2.02 &2.14& \\
		Traj 2  &6.44& 3.91&3.71 &2.08 &2.16& \\
		Traj 3  &6.54& 4.23&3.94 &2.27 &2.45& \\
		Traj 4  &7.25& 4.53&4.14 &3.02 &3.14& \\
		Traj 5  &8.35& 4.95&4.63 &3.34 &3.52& \\
		Traj 6  &8.95& 5.55&5.14 &3.55 &3.82& \\
		\noalign{\smallskip}\hline\noalign{\smallskip}
	\end{tabular}
\end{table}

\subsection{Computational Performance}
To analyze the computational performance of the proposed framework, we determined the average frame processing time across the trajectory sequences. The test platform was a desktop PC with an Intel Xeon E5-1660v3- CPU at 3.00, 8 cores, 32GB of RAM and an NVIDIA Quadro K1200 GPU with 4GB of memory. Execution time tests were done to determine the ratio between the execution time and length of video stitched.  3D reconstruction of 100 frames took 80.54 seconds to process. 3D reconstruction of 200 frames took 180.83 seconds to process. 3D reconstruction of 300 frames took 290.12 seconds to process.That indicates an overall processing time of 919.15 ms per frame pair, implying that the pipeline needs to be accelerated using more effective parallel-computing and GPU power to reach real time performance.

\section*{References}

\bibliography{mybibfile}

\end{document}